\documentclass[11pt]{article}

\usepackage[final]{acl}

\usepackage{times}
\usepackage{latexsym}

\usepackage[T1]{fontenc}

\usepackage[utf8]{inputenc}

\usepackage{microtype}

\usepackage{inconsolata}

\usepackage{graphicx}
\usepackage{amsmath}
\usepackage{array}       
\usepackage{booktabs}
\usepackage{graphicx} 
\usepackage{threeparttable}
\usepackage[table]{xcolor}

\usepackage{multirow}
\usepackage{colortbl}

\usepackage{pgfplots}
\pgfplotsset{compat=1.18} 
\usepackage{tikz} 

\usepackage{listings}
\usepackage{xcolor}
\usepackage{xspace}
\usepackage{float}

\usepackage[most]{tcolorbox}

\definecolor{mygray}{RGB}{230,230,230}
\definecolor{myblue}{RGB}{200,220,240}
\definecolor{mypink}{RGB}{255,200,200}
\newcommand*{\affmark}[1][*]
{}
\newcommand*{\affaddr}[1]{#1}

\newcommand{\dvd}[1]{\textsc{PACE}\xspace}

\title{Finding the Right Evidence: Factor-Guided Coarse-to-Fine \\Reasoning for Long Videos}

\author{Baixuan Xu\affmark[1]\thanks{\quad Equal Contribution},
Yinyui Xu\affmark[1]$^{*}$,
Tianshi Zheng\affmark[1],
Zhaowei Wang\affmark[1],
Weiqi Wang\affmark[1],\\
\textbf{Haochen Shi\affmark[1],
Jiayu Liu\affmark[1],
Qing Zong\affmark[1],
Xiyu Ren\affmark[1],
Xinyu Geng\affmark[1],}\\
\textbf{Zhitao He\affmark[1],
Yangqiu Song\affmark[1]}\\
\affaddr{\affmark[1]Department of Computer Science and Engineering, HKUST, Hong Kong SAR, China}\\
\texttt{bxuan@connect.ust.hk,yqsong@cse.ust.hk}\\}

\begin{document}
\maketitle

\begin{abstract}
While LVLMs rapidly improve, long-video question answering still remains challenging: relevant evidence is sparse, and question-relevant context often fails to provide cues that discriminate the correct answer from plausible alternatives. Diagnostic analysis on a manually annotated subset of MMR-V shows that prior agentic systems substantially improve cue retrieval over direct VLM inference yet fail to achieve a corresponding gain in answer accuracy, indicating that the bottleneck lies in option-discriminative evidence rather than topical relevance alone. We propose \textbf{PACE} (\textbf{P}rogressive \textbf{A}cquisition of \textbf{C}ritical \textbf{E}vidence), a factor-guided framework for long-video evidence acquisition. PACE proceeds in two stages: it first indexes clip-level descriptions guided by question-derived factors without observing the candidate answers; it then uses the candidate answers to derive contrastive cues and queries the index for verification. On MMR-V with the open-source Qwen3-VL backbone, PACE achieves \textbf{42.6\%} accuracy, outperforming direct inference and prior agentic baselines including Deep Video Discovery (DVD). On the same diagnostic subset, PACE recovers \textbf{66.9\%} of the annotated cues, providing empirical evidence that its gains are associated with improved evidence recovery rather than stronger answer-side priors alone. Consistent gains over DVD on \textbf{LVBench}, \textbf{Video-MME}, \textbf{EgoSchema}, and \textbf{LongVideoBench} suggest that option-aware evidence acquisition transfers beyond MMR-V. Code is available at \url{https://github.com/HKUST-KnowComp/PACE}.
\end{abstract}


\begin{figure}[t]
    \centering
    \includegraphics[width=0.9\linewidth]{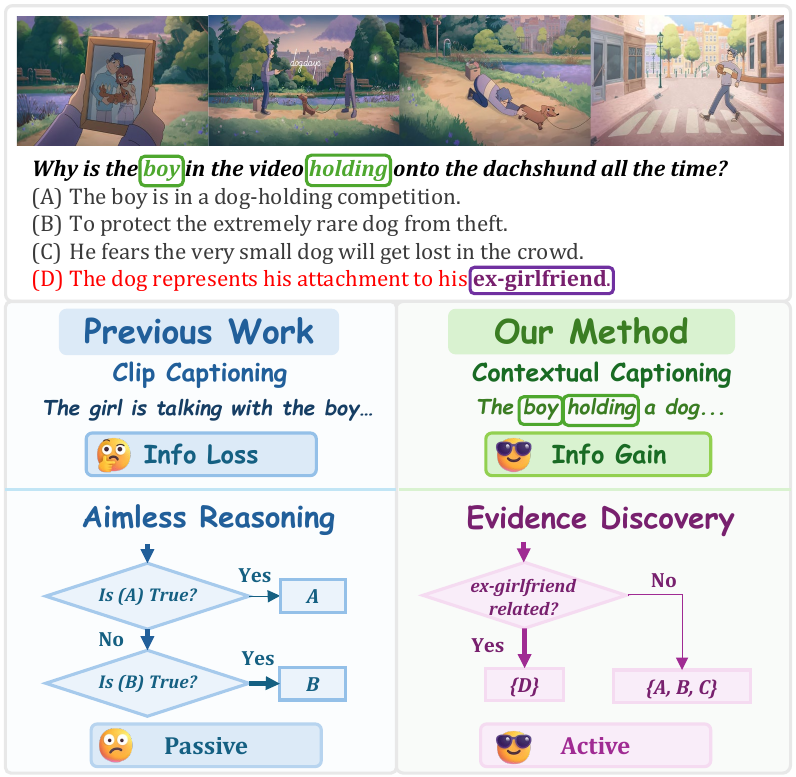}
    \caption{Illustration of the limitation in previous works versus our approach. Left: Previous methods suffer from information loss due to task-agnostic captioning. Right: Our method generates focused, contextual captions to support effective evidence discovery.}
    \label{fig:intro}
    \vspace{-10pt}
\end{figure}

\section{Introduction}

Long-video question answering is often framed as a long-context understanding~\citep{wang2026mmlongbench} problem: as video sequences grow from minutes to hours, a natural assumption is that models benefit from processing more frames, building larger memories, or retrieving more context~\citep{bai2025qwen3vl,wang2026training,wang2024videoagent,zhang2025deep,pang2025mrvideo}. However, access to more context does not necessarily translate into better reasoning. In challenging cases, decision-critical evidence is sparse and visually subtle, surrounded by abundant context that is topically related yet non-decisive. The bottleneck thus lies in the quality of the evidence rather than the abundance of context: such evidence must not only relate to the question but also help discriminate among plausible alternative answers. Yet current pipelines, optimized for question relevance, do not directly target this discriminative property.

Consider the question in Figure~\ref{fig:intro}: \emph{Why is the boy in the video holding onto the dachshund all the time?} Many clips throughout the video show the boy holding or interacting with the dog, and all are clearly relevant to the question. Yet none of them, on their own, distinguishes the four candidate explanations: a dog-holding competition, protection from theft, fear of losing the dog, or attachment to an ex-girlfriend. The decisive cue is a brief shot of a photograph showing the ex-girlfriend with the same dog, which does not match the surface form of the question and is therefore easily overlooked by question-only retrieval. We refer to this failure mode as \emph{option-blind retrieval}: when the retrieval signal is constructed independently of the candidate answers, the system can return abundant question-relevant content while missing the cues that discriminate among them. The challenge is therefore not finding question-relevant clips, but recovering option-discriminative evidence.

To verify that this gap appears in practice, we manually annotate 100 questions from MMR-V~\citep{zhu2025mmrv} with the visual cues human annotators identified as supporting the correct answer, and check whether each cue appears in the systems' inference traces. The most directly comparable open-source agentic baseline, Deep Video Discovery (DVD)~\citep{zhang2025deep}, recovers 56.2\% of these cues, compared with 43.2\% for direct Qwen3-VL inference~\citep{bai2025qwen3vl} and 30.6\% for VideoTree~\citep{Wang2025videotree}. Despite this wide gap in cue recovery, answer accuracy remains comparable across the three systems, with all reaching roughly 54\%. The evidence surfaced in their traces is typically question-relevant but not consistently option-discriminative, indicating that the bottleneck lies in evidence that distinguishes among candidate answers rather than evidence that merely relates to the question. We examine this dissociation in detail in \S\ref{sec:needle_diagnosis}.

We propose \textbf{PACE} (\textbf{P}rogressive \textbf{A}cquisition of \textbf{C}ritical \textbf{E}vidence), a factor-guided two-stage framework for long-video evidence acquisition. The first stage extracts a compact set of question-derived factors covering entities, actions, attributes, and temporal anchors. These factors then guide clip-level video description, which is indexed into an evidence database. This stage proceeds without observing the candidate answers, so the database is not pre-shaped toward any particular hypothesis. The second stage examines the candidate answers, derives the cues that would discriminate among them, and queries the database for matching descriptions to support verification. This write-without-options, read-with-options asymmetry prevents the candidate answers from shaping the indexed evidence, while still allowing them to drive evidence verification.

\label{sec:intro_pace}
We evaluate PACE primarily on MMR-V, where it attains \textbf{42.6\%} accuracy with the Qwen3-VL backbone, compared with 39.5\% for DVD. On the same 100-question diagnostic subset, PACE recovers \textbf{66.9\%} of the annotated cues, substantially higher than 56.2\% for DVD, supporting the view that its accuracy gains are linked to improved evidence recovery rather than answer-side priors alone. PACE also yields consistent gains over DVD on \textbf{LVBench}~\citep{wang2024lvbench}, \textbf{Video-MME}~\citep{fu2024videomme}, \textbf{EgoSchema}~\citep{mangalam2023egoschema}, and \textbf{LongVideoBench}~\citep{wu2024longvideobench}, suggesting that option-aware evidence acquisition transfers beyond the diagnostic MMR-V setting.

Our contributions are summarized as follows:
\begin{itemize}
    \item We identify and quantify \emph{option-blind retrieval} as a failure mode in long-video question answering, using a manually annotated needle-recall diagnostic to show that higher question-relevant cue recovery does not, on its own, translate into higher accuracy.
    \item We propose \textbf{PACE}, a factor-guided two-stage framework that separates question-conditioned evidence indexing from option-aware verification, preventing the candidate answers from shaping the indexed evidence.
    \item We show that PACE consistently improves both answer accuracy and needle recall on MMR-V and four broader long-video benchmarks, with the diagnostic confirming the gains stem from evidence recovery.
\end{itemize}

\section{Related Work}

\subsection{Long Vision-Language Models}

Large vision-language models for long videos have evolved from frame aggregation baselines~\citep{lin2024video,li2025videochat} toward hierarchical compression~\citep{chen2024longvila,li2024videochatflash} and dynamic token pruning~\citep{wang2024retake,wang2025adaretake}. Foundation models such as Qwen-VL~\citep{bai2025qwen3vl,bai2025qwen25,wang2024qwen2}, MMProLong~\citep{wang2026training}, and InternVL/Video~\citep{chen2024internvl,wang2025internvideo25} continue to scale, alongside specialized architectures~\citep{ye2024mplugowl3,zhang2025videollama,chen2024timemarker}.

\begin{figure*}
    \centering
    \includegraphics[width=1\linewidth]{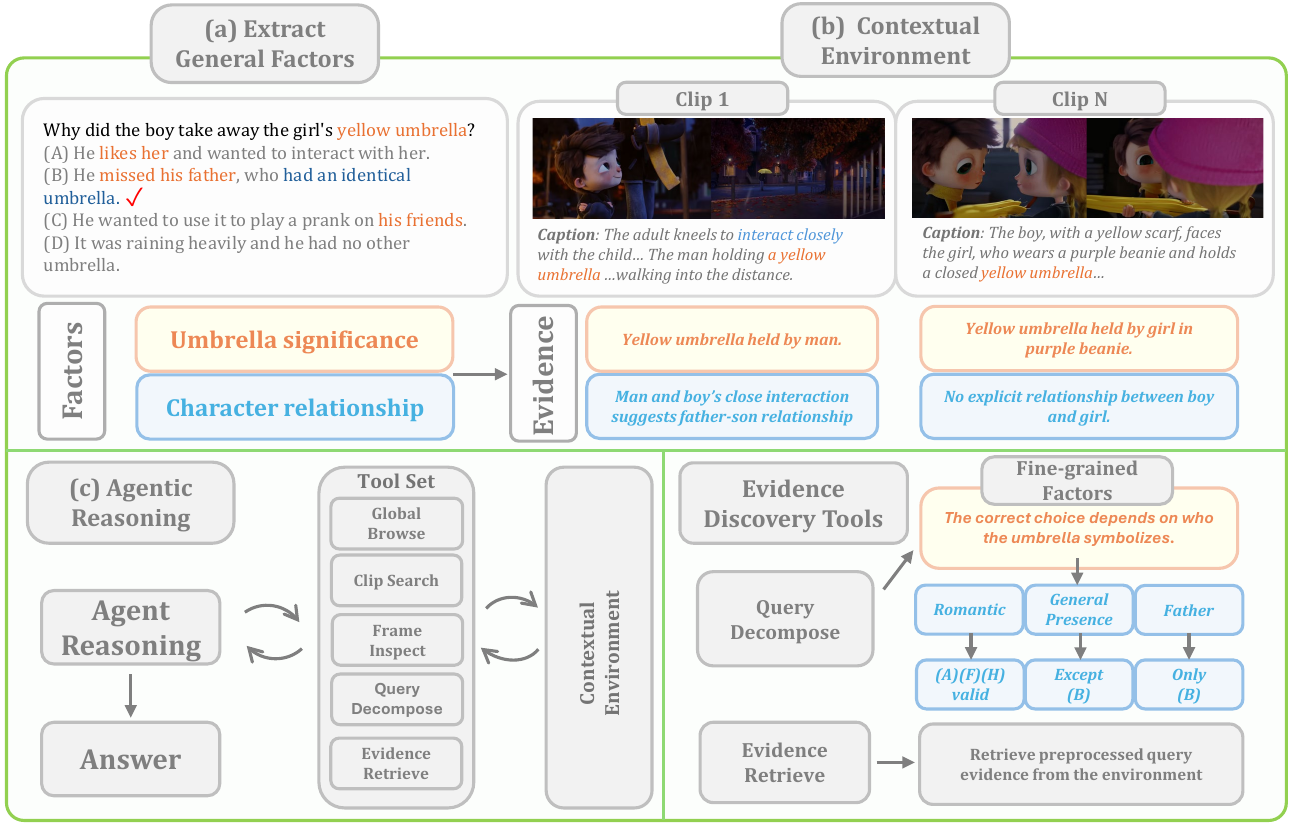}
    \caption{Overview of PACE. The first stage builds an evidence database from clip-level descriptions guided by question-derived factors, without observing the candidate answers. The second stage derives contrastive cues from the candidate answers and queries the database to verify them.}
    \label{fig:framework}
\end{figure*}

\subsection{Agentic Long-Video Reasoning}

Agentic frameworks use LLMs to actively explore video content, evolving from text-based agent designs~\citep{DBLP:conf/emnlp/Xu0S0JFBLYLLYYC24,DBLP:journals/corr/abs-2505-07313,DBLP:journals/corr/abs-2510-07091} toward video-specific instantiations. Recent approaches include retrieval over pre-computed dense captions~\citep{zhang2025deep,pang2025mrvideo}, query-adaptive hierarchies~\citep{Wang2025videotree}, and state-action browsing~\citep{wang2024videoagent,yang2025vca}. Among these, Deep Video Discovery~\citep{zhang2025deep} is the most directly comparable open-source agentic system and serves as our primary point of comparison.
Beyond inference-time video reasoning, recent work has explored scalable agent supervision through grounded GUI trajectories from video, step-level process refinement, and meta-plan optimization~\citep{xiong2026video2gui,xiong-etal-2024-watch,xiong-etal-2025-mpo}; studies of program testing further distinguish code generation from behavioral verification~\citep{xiong-etal-2024-program}, a distinction echoed in PACE's separation of evidence acquisition and answer verification.
While different methods in retrieval mechanisms are designed, these systems do not condition retrieval on the candidate answers. In contrast, our PACE further partitions evidence acquisition along an information boundary, with an indexing stage conditioned on the question and a verification stage that derives contrastive cues from candidate answers, addressing the option-blind retrieval gap quantified in \S\ref{sec:needle_diagnosis}.

\section{The PACE Framework}
We now present PACE, the two-stage framework introduced in \S\ref{sec:intro_pace}. As illustrated in Figure~\ref{fig:framework}, PACE separates evidence acquisition into a question-conditioned indexing stage that runs without observing the candidate answers, and an option-aware verification stage that uses the candidate answers to interrogate the resulting index.

\subsection{Problem Setup}
\label{sec:method_setup}
Let $V=\{v_1,\dots,v_N\}$ denote a long video uniformly partitioned into $N$ non-overlapping clips. Given a question $Q$ and a candidate answer set $\mathcal{A}=\{O_1,\dots,O_n\}$, the task is to select the correct answer $a^\star\in\mathcal{A}$ from evidence acquired over $V$. PACE acquires this evidence in two stages connected by an evidence database $\mathcal{M}$, in which each entry stores a clip identifier, a textual evidence record, and the question-derived factors that the record supports. The two stages differ deliberately in what they observe: stage~1 has access only to $V$ and $Q$, while stage~2 additionally consults $\mathcal{A}$. This write-without-options, read-with-options asymmetry is the central design choice of PACE.

\subsection{Stage 1: Question-Conditioned Evidence Indexing}
\label{sec:method_stage1}
The first stage builds $\mathcal{M}$ entirely from $Q$ and $V$, without observing $\mathcal{A}$. Given $Q$, we prompt a language-capable model to extract a compact set of factors $F_Q=\{f_1,\dots,f_k\}$, where each $f_j$ is a short natural-language descriptor of one informational facet of the question. We allow $F_Q$ to cover entities, actions, relations, attributes, and temporal anchors, and we constrain $|F_Q|$ to a small budget so that the factors remain question-anchored rather than unconstrained scene descriptions. For the example in Figure~\ref{fig:intro}, $F_Q$ contains factors such as \emph{the boy holding the dog}, \emph{interactions between the boy and the dog}, and \emph{earlier scenes featuring the dog}.

The factors then condition clip-level description. For each clip $v_i$, a video-language model produces a textual record that explicitly annotates which factors in $F_Q$ are observed and how. The resulting records are embedded with a sentence encoder and inserted into $\mathcal{M}$, alongside their associated clip identifier and factor tags. Because $\mathcal{A}$ is hidden from this stage, no record in $\mathcal{M}$ is shaped to favor any specific candidate answer, and the same $\mathcal{M}$ serves all options at verification time. We deliberately restrict the factor budget at this stage rather than expanding it: empirically, a small $|F_Q|$ keeps the index focused without over-pruning subtle but discriminative cues, an effect we revisit in \S\ref{sec:factor_impact}.

\subsection{Stage 2: Option-Aware Verification}
\label{sec:method_stage2}
The second stage admits $\mathcal{A}$ for the first time and tests whether the evidence stored in $\mathcal{M}$ distinguishes among the candidates. We instantiate this through a small set of contrastive factors $F_{Q,\mathcal{A}}=\{d_1,\dots,d_m\}$, each of which is a natural-language criterion that distinguishes at least two options in $\mathcal{A}$. The \textsc{QueryDecompose} tool produces $F_{Q,\mathcal{A}}$ from $(Q,\mathcal{A})$ at the start of the trajectory: it consumes the question and the candidate options, returns a short list of contrastive criteria, and emits the option subset that each criterion can rule in or out.

A reasoning agent then drives a state-action-observation loop over $\mathcal{M}$. Its state holds $Q$, $\mathcal{A}$, the current set of unresolved factors in $F_{Q,\mathcal{A}}$, and the textual evidence it has already retrieved. Its action space comprises three tools inherited from prior agentic video frameworks (\textsc{GlobalBrowse}, \textsc{ClipSearch}, \textsc{FrameInspect}; described in Appendix~\ref{app: tool sets}) and two tools specific to PACE: \textsc{QueryDecompose}, used at most once per trajectory to obtain $F_{Q,\mathcal{A}}$, and \textsc{EvidenceRetrieve}, which takes a single contrastive factor as input and returns the top-$k$ records in $\mathcal{M}$ whose factor tags or evidence text match it. \textsc{FrameInspect} is reserved for cases where the textual records leave a contrastive factor unresolved and direct visual inspection of a localized time range is required. The agent terminates when every factor in $F_{Q,\mathcal{A}}$ has been addressed by retrieved or inspected evidence, or when an action budget is exhausted, and then selects a final answer from $\mathcal{A}$.

The two factor sets play complementary roles. $F_Q$ is question-anchored and therefore covers what should be recorded about the video; $F_{Q,\mathcal{A}}$ is option-anchored and therefore covers what should be tested against those records. Treating them as a single object, as a symmetric option-aware retrieval pipeline would, either embeds answer-side priors into $\mathcal{M}$ before any verification can take place, or forces $\mathcal{M}$ to be rebuilt for each option. The asymmetry of PACE avoids both: $\mathcal{M}$ remains independent of $\mathcal{A}$, while verification remains targeted to option-discriminative evidence.

\section{Experiments}
\begin{table*}[!htbp]
\centering
\normalsize
\renewcommand{\arraystretch}{0.95}
\setlength{\tabcolsep}{3.8pt}
\resizebox{0.98\linewidth}{!}{
\begin{tabular}{p{5cm}|m{0.7cm}m{0.7cm}|m{0.7cm}m{0.7cm}|m{0.7cm}|m{0.7cm}|m{0.7cm}|m{0.7cm}|m{0.7cm} |m{0.7cm}|m{0.7cm}|m{0.7cm}}
\toprule
 &  & & \multicolumn{4}{c}{\textbf{Tasks}} & \multicolumn{6}{|c}{\textbf{Video Categories}} \\
\cmidrule(r){2-3} \cmidrule(lr){4-7} \cmidrule(l){8-13}
\textbf{Model} & \multicolumn{2}{c|}{\textbf{Overall}} & \multicolumn{2}{c|}{\textbf{Implicit}} & \multicolumn{2}{c|}{\textbf{Explicit}} & \multicolumn{1}{c|}{\textbf{Art}} & \multicolumn{1}{c|}{\textbf{Life}} & \multicolumn{1}{c|}{\textbf{TV}} & \multicolumn{1}{c|}{\textbf{Film}} & \multicolumn{1}{c|}{\textbf{Ani.}} & \multicolumn{1}{c}{\textbf{Phi.}} \\ 
\midrule
\multicolumn{13}{c}{\emph{Proprietary models}} \\
\textrm{GPT-4o-mini} & \multicolumn{2}{c|}{34.8} & \multicolumn{2}{c|}{38.0} & \multicolumn{2}{c|}{26.3}& 29.5 & 25.4 & 29.6 & 33.0 & 48.7 & 18.6 \\
\textrm{Claude-3.5-Sonnet} & \multicolumn{2}{c|}{43.3} & \multicolumn{2}{c|}{45.0} & \multicolumn{2}{c|}{38.9} & 33.8 & 31.1 & 41.3 & 41.3 & 55.8 & 44.4\\
\textrm{GPT-4o} & \multicolumn{2}{c|}{44.0} & \multicolumn{2}{c|}{46.6} & \multicolumn{2}{c|}{37.6} & 38.1 & 37.3 & 34.9 & 41.0 & 61.6 & 32.6\\
\textrm{GPT-4.1} & \multicolumn{2}{c|}{46.6} & \multicolumn{2}{c|}{49.1} & \multicolumn{2}{c|}{40.3} & 43.2 & 35.6 & 43.9 & 46.5 & 57.1 & 34.9\\
\textrm{Gemini-2.5-Flash} & \multicolumn{2}{c|}{51.2} & \multicolumn{2}{c|}{52.9} & \multicolumn{2}{c|}{46.9} & 45.3 & 39.5 & 50.3 & 47.9 & 65.6 & 34.9 \\
\textrm{o4-mini} & \multicolumn{2}{c|}{52.5} & \multicolumn{2}{c|}{54.6} & \multicolumn{2}{c|}{47.1} & 48.2 & 40.1 & 54.0 & 51.7 & 65.3 & 27.9\\

\midrule 
\multicolumn{13}{c}{\emph{Open-source models}} \\
\textrm{LLaVA-Onevision} & \multicolumn{2}{c|}{6.5} & \multicolumn{2}{c|}{7.0} & \multicolumn{2}{c|}{5.4} & 6.5 & 3.4 &9.5 & 3.8 & 9.8 & 1.2 \\
\textrm{LLaVA-Video} & \multicolumn{2}{c|}{18.4} & \multicolumn{2}{c|}{19.1} & \multicolumn{2}{c|}{15.4} & 14.4 & 11.2 & 13.2 & 17.4 & 21.4 & 12.8 \\
\textrm{Phi-4-multimodal-instruct} & \multicolumn{2}{c|}{26.7} & \multicolumn{2}{c|}{29.4} & \multicolumn{2}{c|}{19.4} & 19.4 & 19.2 & 25.9 & 26.4 & 33.9 & 24.4 \\
\textrm{Cogvlm2-video-llama3}   & \multicolumn{2}{c|}{25.6} & \multicolumn{2}{c|}{25.4} & \multicolumn{2}{c|}{26.1} & 15.5 & 18.3 & 24.7 & 19.1 & 43.2 & 20.8 \\
\textrm{Qwen2.5-VL-7B} & \multicolumn{2}{c|}{30.1} & \multicolumn{2}{c|}{33.7} & \multicolumn{2}{c|}{20.8} & 20.9 & 18.1 & 29.6 & 21.2 & 48.4 & 19.8 \\
\textrm{InternVL2.5-38B} & \multicolumn{2}{c|}{39.9} & \multicolumn{2}{c|}{43.8} & \multicolumn{2}{c|}{29.9} & 30.4 & 28.8 & 30.4 & 37.2 & 57.4 & 29.1 \\
\textrm{Qwen2.5-VL-72B} & \multicolumn{2}{c|}{39.1} & \multicolumn{2}{c|}{41.3} & \multicolumn{2}{c|}{33.4} & 28.9 & 28.2 & 29.1 & 36.5 & 55.6 & 37.2 \\ 
\textrm{Gemma-3-27b-it} & \multicolumn{2}{c|}{42.0} & \multicolumn{2}{c|}{46.5} & \multicolumn{2}{c|}{30.3} & 31.7 & 32.2 & 35.5 & 41.3 & 56.1 & 33.7 \\

\midrule

\multicolumn{13}{c}{\emph{Baseline and Agentic Framework}} \\
\emph{Qwen3-VL-30B-A3B-Thinking} & \multicolumn{2}{c|}{39.3} & \multicolumn{2}{c|}{43.9} & \multicolumn{2}{c|}{27.1} & \multicolumn{1}{c|}{26.6} & \multicolumn{1}{c|}{27.1} & \multicolumn{1}{c|}{34.4} & \multicolumn{1}{c|}{35.9} & \multicolumn{1}{c|}{57.4} & \multicolumn{1}{c}{26.7} \\
\emph{VideoTree} & \multicolumn{2}{c|}{34.4} & \multicolumn{2}{c|}{36.1} & \multicolumn{2}{c|}{30.1} & \multicolumn{1}{c|}{21.1} & \multicolumn{1}{c|}{31.6} & \multicolumn{1}{c|}{29.2} & \multicolumn{1}{c|}{37.5} & \multicolumn{1}{c|}{41.3} & \multicolumn{1}{c}{24.4} \\
\emph{Deep Video Discovery} & \multicolumn{2}{c|}{39.5} & \multicolumn{2}{c|}{42.2} & \multicolumn{2}{c|}{32.6} & \multicolumn{1}{c|}{\textbf{28.8}} & \multicolumn{1}{c|}{21.5} & \multicolumn{1}{c|}{32.3} & \multicolumn{1}{c|}{38.2} & \multicolumn{1}{c|}{58.2} & \multicolumn{1}{c}{\textbf{32.6}} \\
\rowcolor{gray!20}\emph{\dvd} & \multicolumn{2}{c|}{\textbf{42.6}} & \multicolumn{2}{c|}{\textbf{45.3}} & \multicolumn{2}{c|}{\textbf{35.7}} & \multicolumn{1}{c|}{28.1} & \multicolumn{1}{c|}{\textbf{33.3}} & \multicolumn{1}{c|}{\textbf{38.1}} & \multicolumn{1}{c|}{\textbf{39.6}} & \multicolumn{1}{c|}{\textbf{60.8}} & \multicolumn{1}{c}{25.6} \\
\bottomrule
\end{tabular}
}
\caption{Evaluation results (\%) on MMR-V. \textbf{Bold} values indicate the best performance among Baseline and Agentic Framework. The backbone for VideoTree, DVD and \dvd{} is Qwen3-VL-30B-A3B-Thinking.}

\label{tab:mmrv_result}
\end{table*}
We evaluate PACE on MMR-V as our primary diagnostic benchmark and on four broader long-video QA datasets. Beyond overall accuracy, we examine whether the gains arise from genuine evidence recovery via a needle-recall diagnostic, test transfer to broader benchmarks, evaluate robustness across backbone configurations, ablate each stage, and study how the factor budget affects the indexing stage.

\subsection{Setup}
\paragraph{Datasets.} Our primary benchmark is MMR-V~\citep{zhu2025mmrv}, which contains 1,257 questions over 317 videos in six categories and stresses multi-step reasoning under ``multiple needles in a haystack'' conditions. For broader transfer we evaluate on the long-video subset of Video-MME~\citep{fu2024videomme}, the validation set of LVBench~\citep{wang2024lvbench}, the longest-video subset of LongVideoBench~\citep{wu2024longvideobench}, and EgoSchema~\citep{mangalam2023egoschema}. Statistics and preprocessing details are in Appendix~\ref{app:dataset}.

\paragraph{Baselines.} We compare against proprietary VLMs, open-source VLMs, and open-source agentic frameworks; the full lists are in Appendix~\ref{app:models and frameworks}. Our primary point of comparison is Deep Video Discovery (DVD)~\citep{zhang2025deep}. The open-source agentic baselines we evaluate against PACE (DVD throughout, and VideoTree on MMR-V and on the needle-recall diagnostic) are re-implemented under the same Qwen3-VL backbone, video sampling rate, clip duration, and action budget as PACE. Direct Qwen3-VL inference on the needle-recall diagnostic uses the same backbone settings. This matched evaluation isolates accuracy differences from backbone strength, video preprocessing, or inference budget. All other baseline numbers, including non-agentic foundation models, proprietary VLMs, and the broader video-agent comparisons in Appendix~\ref{app:full comparison on general benchmarks}, are quoted from their original benchmark reports.

\paragraph{Implementation.} We use Qwen3-VL-30B-A3B-Thinking~\citep{bai2025qwen3vl} as the unified backbone for both clip-level description and the reasoning agent. Captions and contrastive factors are embedded with \texttt{text-embedding-3-large} following~\citet{zhang2025deep}. Frames are sampled at 2 FPS and resized to 720p, and videos are uniformly partitioned into clips of $T=10$ seconds. The agent runs at most $N=15$ steps, and \textsc{ClipSearch} retrieves the top 16 results by default. The number of question-derived factors $|F_Q|$ is determined dynamically by the LLM under the budget $1 \leq |F_Q| \leq 3$.

\subsection{Main Results on MMR-V}
Table~\ref{tab:mmrv_result} reports performance on MMR-V under the unified Qwen3-VL backbone. PACE attains an overall accuracy of \textbf{42.6\%}, exceeding both VideoTree (34.4\%) and DVD (39.5\%). The gain over DVD is consistent across the implicit-reasoning and explicit-reasoning subsets (+3.1 each), indicating that the improvement is not driven by a single reasoning style.

PACE's largest category-level gains over DVD appear in \emph{Life} (+11.8) and \emph{TV} (+5.8). Videos in these categories typically contain substantial visual noise and loosely structured events, where retrieval pipelines are most prone to surfacing globally relevant but non-discriminative content. The pattern is consistent with our central claim that option-aware verification is most beneficial when the index returns plausible-but-non-decisive context. PACE shows limited or negative gains in \emph{Art} (28.1 vs 28.8) and \emph{Philosophy} (25.6 vs 32.6); questions in these categories often depend on abstract or symbolic interpretation that is only weakly anchored to entity- or action-level cues. We provide an illustrative case study in Appendix~\ref{sec:art_phi_analysis}.

\subsection{Needle-Recall Diagnosis}
\label{sec:needle_diagnosis}
To understand whether PACE's accuracy gains arise from genuine evidence recovery rather than stronger answer-side priors alone, we conduct a needle-recall diagnostic. Two of the authors independently annotate the 100 questions in MMR-V with the minimal set of visual cues required to establish the question premise and to distinguish the correct answer from distractors, and reconcile disagreements through discussion. We then collect the inference traces of four systems (VideoTree, direct Qwen3-VL inference, DVD, and PACE) and use GPT-5 as an LLM-as-judge to determine whether each annotated needle is matched. A needle counts as recovered when the system trace contains visual evidence matching an annotated cue, regardless of whether the final answer is correct; the judge receives the question, the annotated cue, and the retrieved trace evidence, but not the system name. We report needle recall as the fraction of annotated cues recovered per question, averaged over the diagnostic subset. This separates evidence recovery from answer selection, so the recall metric reflects trace quality rather than serving as an accuracy proxy. The full prompt is in Appendix~\ref{app:LLM-as-judge}. The 100 questions function as a probe rather than an exhaustive measurement, and we treat the cross-system gaps as the load-bearing signal.

\begin{figure}[t]
    \centering
    \includegraphics[width=0.95\linewidth]{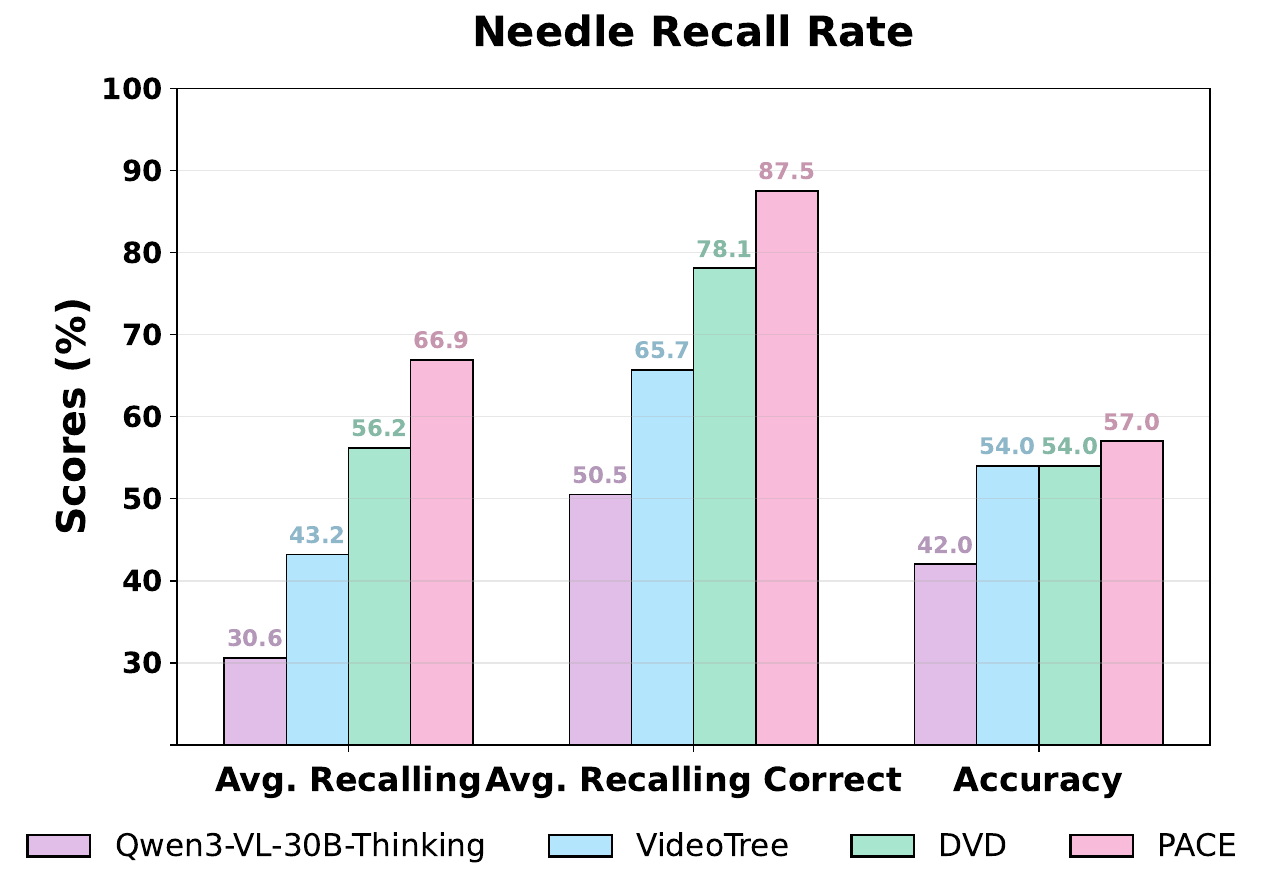}
    \caption{Needle recall and accuracy on the 100-question diagnostic subset of MMR-V. \textbf{Avg.\ Recall} is the overall needle recall rate, \textbf{Avg.\ Recall (Correct)} is the recall rate restricted to correctly answered questions, and \textbf{Accuracy} is the answer accuracy on the subset.}
    \label{fig:needle_recall}
\end{figure}

Figure~\ref{fig:needle_recall} reveals two findings. First, PACE achieves the highest needle recall at \textbf{66.9\%}, well above DVD's 56.2\% and direct Qwen3-VL's 43.2\%, indicating that question-conditioned indexing combined with option-aware verification recovers more decision-supporting evidence than passive scanning or query-agnostic retrieval. Second, retrieval improvement does not by itself translate to accuracy: although DVD raises needle recall by more than 13 points over Qwen3-VL, its answer accuracy on this subset stays at 54.0\%. PACE breaks this pattern, with accuracy at \textbf{57.0\%} alongside the recall gain. The dissociation between recall and accuracy in DVD, and its absence in PACE, is consistent with the asymmetry design: surfacing question-relevant content is necessary but not sufficient, and the option-aware verification stage is what converts retrieved evidence into decision-relevant signal.

We highlight two interpretive caveats. First, recall and accuracy are measured separately by design: recall captures whether decision-supporting evidence appears anywhere in the inference trace, while accuracy captures whether the agent commits to the correct answer at the end. The two dissociate (as DVD shows) when the agent surfaces relevant evidence but cannot use it to discriminate among candidates, and align (as PACE shows) when option-aware verification connects evidence to the decision. Second, the absolute recall numbers may shift under different judge models or annotation conventions; we therefore treat the cross-system gap, not the absolute level, as the load-bearing signal. PACE's relative position above DVD is consistent across the implicit and explicit reasoning splits of the diagnostic subset, suggesting that the gain is not specific to a particular reasoning style.

\subsection{Transfer to Broader Long-Video Benchmarks}
\begin{table}[t]
    \centering
    \small
    \setlength{\tabcolsep}{4.2pt}
    \renewcommand\arraystretch{1.8}
    \resizebox{\linewidth}{!}{
    \begin{tabular}{l cccc}
        \toprule
        \multirow{2}{*}{Method} 
        & LVBench & LongVideoBench & Video-MME & EgoSchema \\
        & Overall & Val (Long) & Long (w/o sub) & Val \\
        \midrule
        Qwen3-VL-30B-A3B-Thinking 
            & 59.2 & 46.1 & 59.4 & 66.0 \\
        Deep Video Discovery 
            & 62.3 & 54.6 & 57.2 & 67.9 \\
        \rowcolor{gray!20}
        \textbf{PACE} 
            & \textbf{62.6} & \textbf{55.1} & 58.9 & \textbf{68.1} \\
        \bottomrule
    \end{tabular}
    }
    \caption{
    Controlled transfer results on general long-video benchmarks using the Qwen3-VL-30B backbone.
    }
    \label{tab:long_video_benchmarks}
\end{table}
We next test whether PACE's evidence-recovery gains transfer beyond MMR-V. Table~\ref{tab:long_video_benchmarks} reports controlled transfer results on four broader long-video QA benchmarks under the same Qwen3-VL-30B backbone. PACE improves over DVD by +0.3 on LVBench, +0.5 on LongVideoBench, +0.2 on EgoSchema, and +1.7 on Video-MME, improving on all four benchmarks. The largest gain appears on Video-MME, while the smaller margins on LVBench, LongVideoBench, and EgoSchema suggest that option-aware verification adds most value when question relevance alone does not distinguish the options. Full comparisons against additional video-agent and foundation-model baselines are in Appendix~\ref{app:full comparison on general benchmarks}.

\subsection{Backbone Robustness}
\begin{table}[t]
\centering
\renewcommand\arraystretch{1.1}
\resizebox{\linewidth}{!}{
\begin{tabular}{lccc}


\toprule
\textbf{Backbone Model} & \textbf{DVD} & \textbf{PACE} & \textbf{Delta} \\
\midrule

Qwen2.5-VL-7B & 34.2 & \textbf{37.6} & +3.4 \\
Qwen3-VL-8B   & 34.0 & \textbf{36.1} & +2.1 \\
Qwen3-VL-30B-A3B-Thinking & 39.5 & \textbf{42.6} & +3.1 \\
\bottomrule

\end{tabular}
}
\caption{Performance comparison (\%) between DVD and PACE frameworks on the MMR-V benchmark using different backbone configurations. Delta shows the absolute improvement of PACE over DVD. For Qwen2.5-VL-7B, the reasoning agent is Qwen3-8B.}
\label{tab:backbone_ablation}
\end{table}
Beyond varying the benchmark, we test whether PACE's gain depends on a specific Qwen3-VL configuration by evaluating it on additional backbone setups. Table~\ref{tab:backbone_ablation} reports accuracy on a composite setup (Qwen2.5-VL-7B for vision with Qwen3-8B for reasoning), a unified smaller backbone (Qwen3-VL-8B), and the primary 30B configuration. PACE outperforms DVD by 3.4, 2.1, and 3.1 absolute points respectively, with the same direction of improvement in every case. The smaller absolute gain on the 8B unified backbone is consistent with weaker base captioning quality at that scale: when the indexing stage cannot record question-anchored facets reliably, the verification stage has less to query against. Because evaluation is deterministic under fixed decoding, we report robustness across model configurations rather than seed variance, and the consistency of the gain across three backbone configurations suggests that the gain is structural rather than tied to a particular model scale or architecture.

\subsection{Ablation Studies}
\begin{table}
  \centering
  \begin{tabular}{lc}
    \hline
    \textbf{Model/Agent} & \textbf{Accuracy} \\
    \hline
    \dvd{}                         & \textbf{42.6}\ \\
    -\textbf{w/o} question-conditioned indexing  & $41.8~_{\textcolor{gray}{-0.8\downarrow}}$ \\
    -\textbf{w/o} option-aware verification  & $41.5~_{\textcolor{gray}{-1.1\downarrow}}$ \\
  \hline
  \end{tabular}
  \caption{Ablation study of \dvd{} on MMR-V. We replace question-conditioned captions with standard captions or remove option-aware verification.}
  \label{tab:ablation}
\end{table}
Table~\ref{tab:ablation} ablates each stage of PACE on MMR-V. Recall that stage 1 builds the evidence database $\mathcal{M}$ from question-derived factors $F_Q$ without observing $\mathcal{A}$, and stage 2 derives contrastive factors $F_{Q,\mathcal{A}}$ from the candidate answers and queries $\mathcal{M}$ for matching descriptions. The two ablation variants below remove different sides of this asymmetry; both retain full agent access to $\mathcal{A}$ at answer commitment.

\paragraph{Removing question-conditioned indexing.} Replacing question-conditioned clip-level descriptions with standard captions reduces accuracy to 41.8\% (-0.8). Without conditioning the index on $F_Q$, the database lacks the question-anchored facets that the verification stage queries against.

\paragraph{Removing option-aware verification.} Disabling stage 2 (no \textsc{QueryDecompose} call, no contrastive-factor retrieval) reduces accuracy from 42.6\% to 41.5\% (-1.1). Without the option-aware step, the agent operates on a question-only index and cannot isolate the option-discriminative cues that distinguish the correct answer from distractors.

\paragraph{Joint effect.} The two stage-level losses sum to 1.9 points relative to the full PACE (42.6\%), while the gap to DVD on the same benchmark is 3.1 points; the two stages therefore appear complementary rather than independently additive. The indexing-only configuration (41.5\%) records useful facets without resolving option support, while the verification-only configuration (41.8\%) issues contrastive queries against an index that lacks the question-anchored facets needed for contrastive retrieval.

\subsection{Factor Impact on the Indexing Stage}
\label{sec:factor_impact}

\begin{figure}
    \centering
    \includegraphics[width=0.9\linewidth]{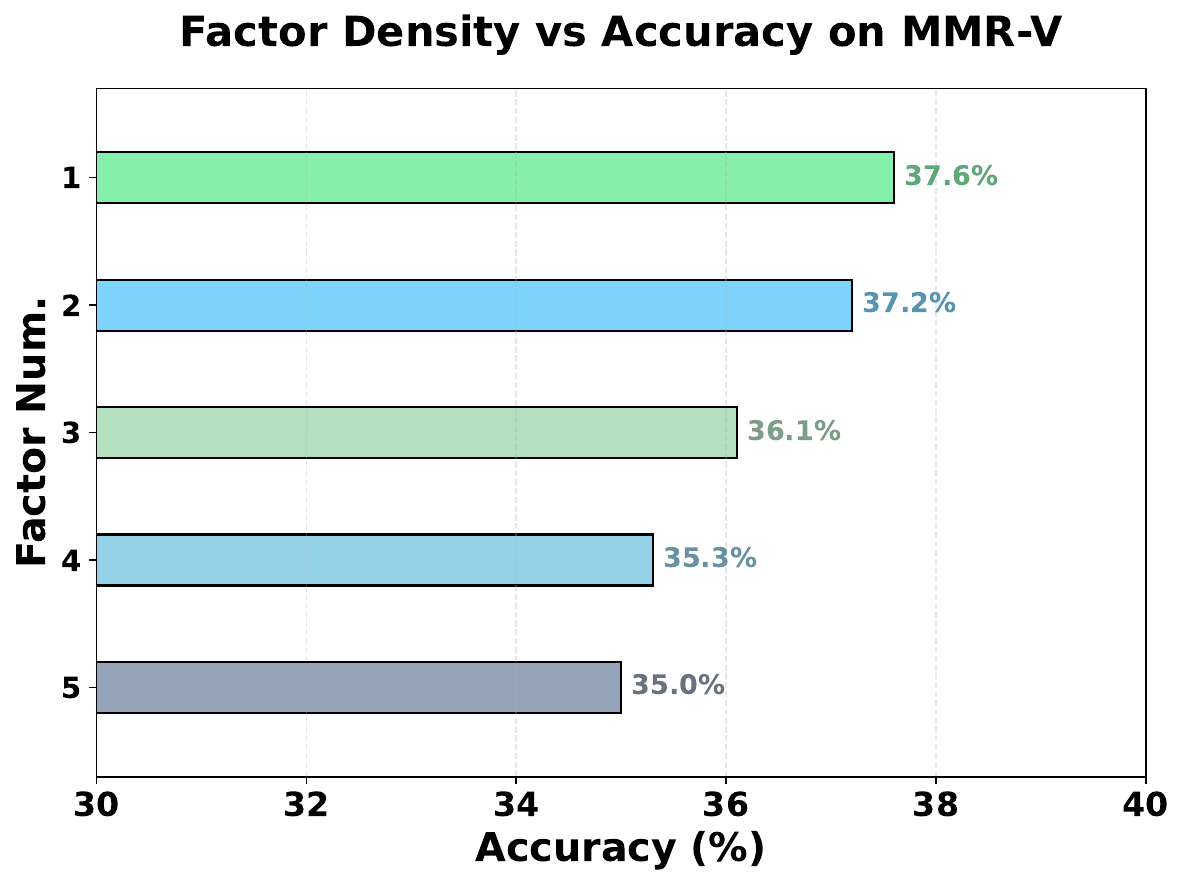}
    \caption{Impact of the number of coarse query-conditioned factors on MMR-V accuracy.}
    \label{fig:needle_density}
\end{figure}
Finally, we examine how the factor budget $|F_Q|$ affects the indexing stage. Under the indexing-stage-only setting, identical to the \emph{w/o option-aware verification} row in Table~\ref{tab:ablation}, we vary the number of forced factors $|F_Q|$ from 1 to 5. As shown in Figure~\ref{fig:needle_density}, accuracy peaks at $|F_Q|=1$ or $2$, drops to 36.1\% at $|F_Q|=3$, and to 35.0\% at $|F_Q|=5$. Under the full PACE pipeline with dynamic $|F_Q| \in [1,3]$, the LLM concentrates extraction in the 1--2 factor regime, consistent with the 41.5\% indexing-only result.

The drop at higher factor counts reflects a packing trade-off. Each additional factor narrows what the captioner attends to within a fixed context window, prematurely filtering cues that the verification stage would later need. Keeping $|F_Q|$ compact preserves question-anchored facets without over-pruning the index, and delegates the burden of fine-grained selection to verification rather than to indexing. A complementary ablation on clip duration (Appendix~\ref{sec:appendix_additional_ablations}) shows the same pattern: $T=10$s outperforms 5s and 20s alternatives, indicating that indexing benefits from moderate granularity rather than denser or coarser segmentation.

\subsection{Tool-Use Dynamics}
\label{sec:tool_use_analysis}

\begin{figure}
    \centering
    \includegraphics[width=0.9\linewidth]{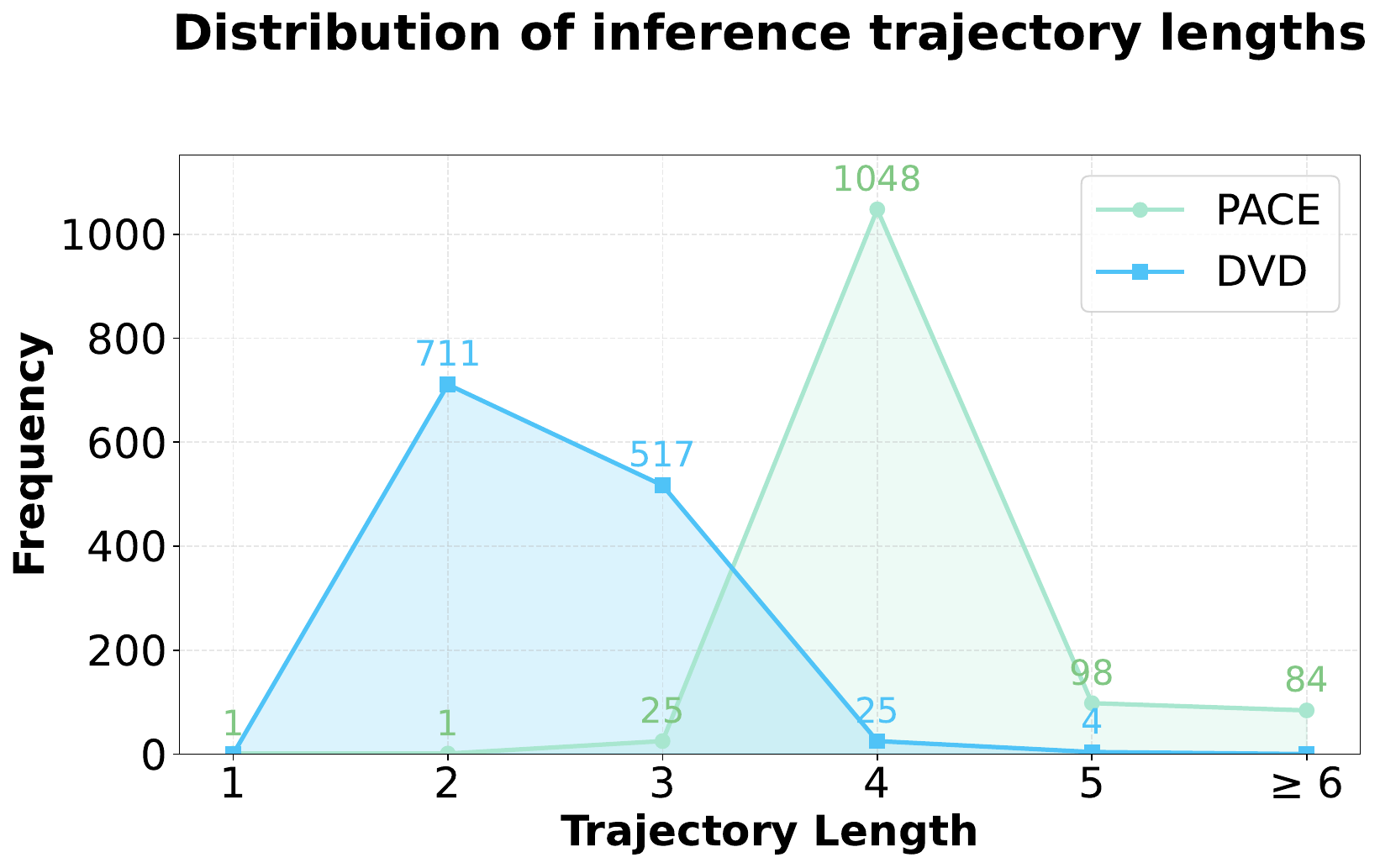}
    \caption{Distribution of inference trajectory lengths of DVD and our proposed \dvd{} framework.}
    \label{fig:num_call_tools}
\end{figure}
The factor-impact analysis suggests that PACE shifts the burden of fine-grained selection from indexing to verification. To check whether this shift is reflected at the tool-use level, we analyze how PACE allocates its action budget on MMR-V. As shown in Figure~\ref{fig:num_call_tools}, most episodes ($83.4\%$) terminate in exactly four steps, converging to a stable three-action sequence followed by answer commitment: the agent first calls \textsc{GlobalBrowse} to ground the question in the full video ($0.99$ calls/episode on average), then \textsc{QueryDecompose} to extract contrastive factors ($0.99$), and finally one or two \textsc{FrameInspect} calls ($1.08$) to verify the most discriminative cues. Mid-level retrieval tools are used sparingly: \textsc{ClipSearch} averages $0.05$ calls/episode, and \textsc{EvidenceRetrieve} averages $0.11$. To check whether these tools are actually load-bearing on the queries that invoke them, we run a conditional ablation: for each tool, we identify the subset of MMR-V questions where PACE invokes it during the original run, and re-evaluate PACE on that subset with the tool removed. As shown in Table~\ref{tab:low_usage_ablation}, removing \textsc{EvidenceRetrieve} drops accuracy by 7.7 points (38.4\% to 30.7\%) on its 65 questions, and removing \textsc{ClipSearch} drops accuracy by 5.4 points (46.4\% to 41.0\%) on its 56 questions. These tools therefore carry weight on the questions where the agent uses them, even though their average usage across all questions is low.
\begin{table}[t]
\centering
\renewcommand\arraystretch{1.1}
\resizebox{\linewidth}{!}{
\begin{tabular}{lcccc}
\toprule
\textbf{Tool} & \textbf{w/ Tool (\%)} & \textbf{w/o Tool (\%)} & \textbf{Drop} & \textbf{Samples} \\
\midrule
\textsc{EvidenceRetrieve} & 38.4 & 30.7 & $-$7.7 & 65 \\
\textsc{ClipSearch}       & 46.4 & 41.0 & $-$5.4 & 56 \\
\bottomrule
\end{tabular}
}
\caption{Conditional ablation study evaluating the performance on the specific subsets of samples where low-usage tools were originally invoked. The notable performance drops indicate the necessity of these tools when explicitly required.}
\label{tab:low_usage_ablation}
\end{table}

\subsection{Token Cost}
\label{sec:token_cost}
\begin{table}[t]
\centering
\renewcommand\arraystretch{1.1}
\resizebox{\linewidth}{!}{
\begin{tabular}{lcc}
\toprule
\textbf{Stage} & \textbf{Avg. Prompt Token} & \textbf{Avg. Completion Token} \\
\midrule
Captioning (per clip)  & 17,584.6 & 1,109.5 \\
Reasoning (per query)  & 68,164.0 & 10,262.0 \\
\bottomrule
\end{tabular}
}
\caption{Average prompt and completion token counts during the captioning and reasoning stages.}
\label{tab:token_counts}
\end{table}
PACE consumes on average $17{,}584$ prompt and $1{,}109$ completion tokens per clip during indexing, and $68{,}164$ prompt and $10{,}262$ completion tokens per query during reasoning (Table~\ref{tab:token_counts}). The reasoning phase dominates the per-query cost. Indexing cost scales linearly with video length, while the reasoning action count is capped by the agent's $N=15$ action budget.

\section{Conclusion}
We propose PACE, a factor-guided two-stage framework that addresses the option-blind retrieval bottleneck in long-video question answering through a write-without-options, read-with-options asymmetry: an indexing stage builds an evidence database from question-derived factors before observing the candidate answers, and a verification stage queries the index using contrastive cues derived from the candidates. PACE attains 42.6\% accuracy and 66.9\% needle recall on MMR-V, with consistent gains across four broader long-video benchmarks and three backbone configurations.

\section*{Limitations}
While PACE demonstrates strong performance in long-video understanding, it still has several limitations. First, compared with end-to-end models, PACE incurs higher inference latency and computational cost because it relies on multiple rounds of model interaction, including factor decomposition and tool invocation (see \S\ref{sec:token_cost} for the per-stage token budget). Although this deliberative process is important for reducing hallucinations in complex scenarios, it also increases time-to-solution. Future work may explore distilling these reasoning trajectories into lighter-weight models to improve efficiency.

Second, PACE mainly improves evidence acquisition. Its gains therefore concentrate on tasks where success depends on locating and composing sparse, decision-critical evidence, and may be smaller when the main bottleneck shifts to abstract semantic, symbolic, or thematic interpretation. In particular, aggressively narrowing the context around query-relevant factors may inadvertently filter out subtle but important cues that are only weakly aligned with the surface form of the query.

Third, PACE still depends on the perception and instruction-following ability of the underlying VLM. Errors caused by severe visual ambiguity, OCR failures, or weak visual grounding in the backbone model may propagate through the agentic reasoning process. As stronger open-source VLMs become available, we expect PACE to benefit from these improvements as well.

Finally, our current implementation is evaluated primarily in multiple-choice settings, where the candidate answer space is explicitly provided. This makes option-aware evidence discrimination directly applicable, but also leaves open the question of how to instantiate the hypothesis space in open-ended QA. One natural extension is a draft-then-verify pipeline: a generative model first proposes a small set of plausible candidate answers or interpretations, which are then treated as an implicit hypothesis space for PACE's contrastive evidence verification. We leave a systematic evaluation of this open-ended extension to future work.


\section*{Ethics Statement}
Data Usage and Privacy: This work utilizes publicly available video understanding benchmarks (MMR-V, LVBench, VideoMME, EgoSchema). We have adhered to the usage licenses and terms of service for all datasets. No new personally identifiable information or private video data was collected or annotated involving human subjects for the purpose of this study.

Bias and Safety: Our framework, PACE, operates on top of pre-trained Vision-Language Models.
Consequently, it may inherit social biases or stereotypes present in the backbone model's training data. While our ``coarse-to-fine'' reasoning strategy is designed to mitigate hallucinations\textemdash thereby reducing the generation of factually incorrect or misleading content\textemdash we acknowledge that the underlying model may still exhibit biased behaviors in open-ended generation. 
We advise users to exercise caution and implement safety filtering when deploying such agentic systems in real-world scenarios.

Computational Impact: We acknowledge that agentic frameworks involving iterative tool use and multi-turn inference consume more computational resources (and thus energy) than single-pass models. We believe this cost is justified by the significant improvements in reasoning reliability for complex long-form video tasks.

Potential Misuse: Advanced video understanding capabilities could theoretically be repurposed for unauthorized surveillance or privacy intrusion. We condemn such misuse and emphasize that PACE is developed strictly for assisting in information retrieval and enhancing the accessibility of video content.
\section*{Acknowledgments}
The authors of this paper were supported by the National Key Research and Development Program of China (2025YFE0200500), the ITSP Platform Research Project (ITS/189/23FP) from ITC of Hong Kong, SAR, China, and the AoE (AoE/E-601/24-N), the CRF (No. C6004-25G), the RIF (R6021-20) and the GRF (16205322) from RGC of Hong Kong, SAR, China.


\bibliography{custom}

\clearpage
\appendix

\section{Tool Sets}
\label{app: tool sets}
\begin{description}
    \item[Global Browse:] Establishes high-level context by returning two types of global summaries: a pre-constructed \textit{subject-centric} summary (derived during dataset construction) and a query-dependent \textit{event-centric} summary. The latter is generated on-the-fly by instructing a VLM to identify noteworthy events relevant to the user query from uniformly sampled frames.

    \item[Clip Search:] Enables mid-level exploration via dense vector retrieval. By calculating the cosine similarity between an agent-synthesized query $\hat{Q}$ and pre-computed clip caption embeddings, this tool retrieves the top-$k$ relevant clips with timestamps. This mechanism supports an iterative ``chain-of-query'' approach, allowing the agent to progressively refine temporal constraints and verify context.

    \item[Frame Inspect:] Facilitates fine-grained visual analysis within a specific temporal range $[t_s, t_e]$. This tool processes raw frames (capped at 50 samples for efficiency) using an open-ended VQA prompt defined by the agent. It is designed to extract subtle evidence—such as object counts, attributes, or spatial relationships—that is often omitted in high-level captions or summaries.
\end{description}

\section{Detail of Dataset}
\label{app:dataset}
In this section, we provide an overview of the benchmarks employed to evaluate capabilities in long-video understanding:

\begin{itemize}
    \item \textbf{Video-MME} \citep{fu2024videomme} is a comprehensive evaluation benchmark encompassing a diverse array of video types with durations ranging from 11 seconds to 1 hour. In our experimental setup, we evaluate the ``Long'' subset without subtitles, which consists of 300 videos and 900 questions.
    
    \item \textbf{LVBench} \citep{wang2024lvbench} focuses on "extreme" long-video understanding, designed to test a model's long-term memory and information extraction capabilities over content spanning several hours. We evaluate our framework on the full benchmark, which comprises 1,549 questions across 103 videos.
    
    \item \textbf{LongVideoBench} \citep{wu2024longvideobench} emphasizes long-context referring reasoning, featuring 6,678 multiple-choice questions based on 3,763 videos. For this study, we evaluate the validation subset within the $(900\text{s}, 3600\text{s}]$ duration range, totaling 564 questions and 188 videos.
    
    \item \textbf{EgoSchema} \citep{mangalam2023egoschema} serves as a diagnostic framework for further investigating long-video comprehension. The model's performance is assessed against the validation set, which consists of 500 videos and an equivalent number of associated questions.
\end{itemize}

\section{Models and Frameworks}
\label{app:models and frameworks}
In this section, we list the models adopted as baselines on MMR-V and other general video benchmarks.
For proprietary models, we select: GPT-4o-mini-2024-07-18~\citep{hurst2024gpt}, Claude-3.5-Sonnet-20241022, GPT-4o-2024-11-20, GPT-4.1-2025-04-14, Gemini-2.5-Flash~\citep{DBLP:journals/corr/abs-2507-06261}, o4-mini-2025-04-16, Gemini-2.0-Flash~\citep{DBLP:journals/corr/abs-2312-11805}, OpenAI-o3, Seed-1.8.

For open-source models: LLaVA-Onevision~\citep{DBLP:journals/tmlr/0080ZGZ00ZZL0L25}, LLaVA-Video~\citep{DBLP:journals/tmlr/ZhangWLLMLL25}, Phi-4-multimodal-instruct, Cogvlm2-video-llama3~\citep{DBLP:journals/corr/abs-2408-16500}, Qwen2.5-VL-7B~\citep{bai2025qwen25}, InternVL2.5-38B~\citep{DBLP:journals/corr/abs-2412-05271}, Qwen2.5-VL-72B, Gemma-3-27b-it, mPLUG-Owl3, InternVL2.5-78B, Qwen-3-VL-30B-A3B, Qwen-3-VL-235B-A22B~\citep{bai2025qwen3vl}.

For agentic frameworks: VideoTree~\citep{Wang2025videotree}, VideoAgent, VCA~\citep{yang2025vca}, MR.Video, Deep Video Discovery.

\section{Full Comparison on General Benchmarks}
\label{app:full comparison on general benchmarks}
In this section, we provide a comprehensive performance comparison of PACE against additional foundation models and video-agent frameworks on broader long-video benchmarks. 
The detailed evaluation results are reported in Table~\ref{tab:appendix_long_video_mid}.
\begin{table}[h]
    \centering
    \small
    \setlength{\tabcolsep}{4.5pt}
    \renewcommand\arraystretch{1.15}
    \resizebox{\linewidth}{!}{
    \begin{threeparttable}
        \begin{tabular}{l cccc}
            \toprule
            \multirow{2}{*}{Method} 
            & LVBench & LongVideoBench & Video-MME & EgoSchema \\
            & Overall & Val (Long) & Long (w/o sub) & Val \\
            \midrule
            \multicolumn{5}{l}{\textcolor{gray}{\textit{Proprietary foundation models}}} \\
            Gemini-2.0-Flash & 48.3 & 45.7 & 63.0 & 71.2 \\
            GPT-4o           & 48.9 & 60.9 & 65.3 & 70.4 \\
            OpenAI o3        & 57.1 & 60.6 & 64.7 & 63.2 \\
            Seed-1.8         & 73.0 & 77.4 & 87.8 & -- \\
            \midrule
            \multicolumn{5}{l}{\textcolor{gray}{\textit{Open-source foundation models}}} \\
            mPLUG-Owl3         & 43.5 & -- & 50.1 & -- \\
            InternVL2.5-78B    & 43.6 & -- & 62.6 & -- \\
            Qwen2.5-VL-72B     & 47.7 & -- & 63.9 & -- \\
            Qwen3-VL-235B-A22B & 63.6 & -- & 79.0 & -- \\
            \midrule
            \multicolumn{5}{l}{\textcolor{gray}{\textit{video-agent frameworks}}} \\
            VideoTree  & 28.8 & --   & --   & 67.0 \\
            VideoAgent & 29.3 & --   & --   & 63.2 \\
            VCA        & 41.3 & --   & --   & 73.6 \\
            MR. Video  & 60.8 & 61.6 & 61.8 & 73.0 \\
            \midrule
            \multicolumn{5}{l}{\textcolor{gray}{\textit{Controlled Qwen3-VL-30B setting}}} \\
            Qwen3-VL-30B-A3B-Thinking & 59.2 & 46.1 & 59.4 & 66.0 \\
            Deep Video Discovery       & 62.3 & 54.6 & 57.2 & 67.9 \\
            \rowcolor{gray!20}
            \textbf{PACE}              & 62.6 & 55.1 & 58.9 & 68.1 \\
            \bottomrule
        \end{tabular}
        \begin{tablenotes}[flushleft]
            \footnotesize
            \item This table provides a full contextual comparison. Published foundation-model and video-agent results are collected from their respective papers when available. The controlled comparison in the bottom block uses Qwen3-VL-30B-A3B-Thinking as the backbone.
        \end{tablenotes}
    \end{threeparttable}
    }
    \caption{Full comparison on general long-video understanding benchmarks.}
    \label{tab:appendix_long_video_mid}
\end{table}

\section{Analysis of Limitations in Art and Philosophy Categories}
\label{sec:art_phi_analysis}
As briefly noted in the main text, while PACE excels at retrieving concrete information, we observe a relative performance limitation in categories such as \textbf{Art} and \textbf{Philosophy}. 
This underperformance is primarily related to the relatively abstract and metaphor-driven nature of questions in these domains, which can sometimes conflict with our framework's emphasis on extracting concrete, entity-centric factual cues.

Specifically, our approach encourages the VLM to decompose queries into entity- and event-centric factors. 
This is highly beneficial when answers depend on identifiable factual cues in long videos.
However, Art and Philosophy questions are often less anchored to explicit entities and more dependent on long-range themes, symbolism, or metaphorical connections. 

As a result, the retrieved captions, which are biased toward concrete, decomposed factual signals, may inadvertently steer the model toward surface-level evidence and away from the intended abstract reasoning, leading to degraded performance.
This limitation is clearly illustrated in our case study on metaphor understanding (Table~\ref{tab:case study metaphor}).
When asked to interpret the symbolism of a man looking into a mirror, the agent’s tools extract literal, surface-level visual connections (e.g., the physical reflection of a child) rather than the deeper psychological themes. 
Guided by this concrete visual verification, the agent incorrectly selects an option based on literal life stages and time passage, failing to grasp the intended abstract introspection.



\section{Prompt for LLM-as-Judge}
\label{app:LLM-as-judge}
\begin{tcolorbox}[
    breakable,
    width=\columnwidth,
    colback=white,
    colframe=gray,
    title=Prompt for LLM-as-Judge,
    arc=0mm,
    boxrule=0.5pt,
    left=2pt, right=2pt, top=2pt, bottom=2pt,
    before skip=5pt, after skip=5pt,
    fonttitle=\bfseries\small,
]
\begin{lstlisting}[
    basicstyle=\ttfamily\footnotesize,
    numbers=none,
    tabsize=2,
    breaklines=true,
    breakatwhitespace=true,
    aboveskip=2pt,
    belowskip=2pt,
]
// system prompt
You are an evaluator for video understanding agents. Your task is to analyze an agent's reasoning trace against a set of predefined, gold-standard information points called "needles." For each needle, determine if the agent **successfully retrieved and considered** the relevant information based on its own reasoning and tool outputs.

// user prompt
**Input Structure:**
You will be given:
1.  **`question_data`**: A JSON object containing:
    *   `question_idx`: Identifier.
    *   `question.content`: The main question.
    *   `question.needles`: A list of needles required to answer the question itself.
    *   `options`: A list of answer choices.
    *   For each option in `options`:
        *   `content`: The option text.
        *   `needles`: A list of needles that are **specifically relevant to confirming or rejecting this particular option**. Each needle has:
            *   `key_point`: The aspect to check.
            *   `answer`: The ground-truth answer (Yes/No or a specific fact).
            *   `description`: A detailed explanation.
2.  **`agent_trace`**: A list of messages (assistant reasoning and tool responses) representing the agent's complete process to answer the question.

**Your Task:**
Analyze the provided `agent_trace`. For **EVERY needle** listed under `question.needles` and under *each* option's `needles`, determine if the information contained in that needle was **retrieved by the agent**.

**Criteria for "Retrieved":**
A needle is considered **`retrieved`** if the agent's **`reasoning_content`** or the **`content` from any `tool` response** explicitly mentions, strongly implies, or logically utilizes the fact described in the needle's `answer` or `description`. Look for semantic equivalence, not exact word matching.
*   If the information is present and consistent with the needle, mark it as `retrieved`.
*   If the information is **absent** or the agent/tool shows **no awareness** of it, mark it as `not_retrieved`.
*   If the agent/tool asserts something that **directly contradicts** the needle's `answer`, mark it as `contradicted`.```

**Output Format:**
You MUST output **ONLY** a valid JSON object adhering to the following schema. No other text, explanation, or markdown.

```json
{
  "question_idx": <integer from input>,
  "final_answer": "<The final answer chosen by the agent, e.g., (B)>",
  "analysis": {
    "question_needles": [
      {
        "key_point": "<string from input>",
        "status": "retrieved | not_retrieved | contradicted",
        "evidence": "<Brief quote or summary from agent_trace that supports the status. Use 'N/A' for not_retrieved.>"
      }
    ],
    "option_analysis": [
      {
        "option_content": "<string, e.g., (A) Difficulties in life.>",
        "needles": [
          {
            "key_point": "<string from input>",
            "status": "retrieved | not_retrieved | contradicted",
            "evidence": "<Brief quote or summary from agent_trace that supports the status. Use 'N/A' for not_retrieved.>"
          }
        ]
      }
    ]
  },
  "summary": {
    "total_needles_count": <integer, total needles analyzed (question + all options)>,
    "retrieved_count": <integer>,
    "contradicted_count": <integer>,
    "not_retrieved_count": <integer>
  }
}
```

**Analysis Instructions:**
1.  Carefully read the entire `agent_trace`. Note the agent's final answer.
2.  For each needle under `question.needles`:
    *   Search the trace for any mention or logical use of the information described in `answer`/`description`.
    *   Assign a `status` and provide `evidence`.
3.  For each `option` in the list:
    *   For each needle under that option's `needles`:
        *   Search the trace. Does the agent recognize the fact needed to judge this specific option? (e.g., for a needle with `answer: "No"`, does the agent note the absence of that feature?)
        *   Assign a `status` and provide `evidence`.
4.  Calculate the summary counts based on all needle statuses.

**Now, analyze the following data:**

Question Data:
```json
QUESTION_DATA
```

Agent Trace:
```json
TRACE_JSON
```
\end{lstlisting}
\end{tcolorbox}
\label{tab:prompts judge}
\captionof{table}{The prompts for evaluating the needle recalling rate.}

\section{Impact of Clip Duration}
\label{sec:appendix_additional_ablations}

We evaluate the clip duration ($T$) used for environment construction in Table \ref{tab:clip_duration}.
Setting $T = 10s$ yields optimal accuracy (42.6\%), outperforming 5s (37.2\%) and 20s (38.1\%).
This highlights 10s as the ideal trade-off: 5s clips fragment the context and increase retrieval noise, whereas 20s clips produce overly coarse captions that reduce the temporal specificity crucial for effective evidence discovery.
\begin{table}[ht]
\centering
\begin{tabular}{lc}
\toprule
\textbf{Clip Duration ($T$)} & \textbf{Accuracy (\%)} \\
\midrule
5s  & 37.2 \\
10s & \textbf{42.6} \\
20s & 38.1 \\
\bottomrule
\end{tabular}
\caption{Ablation study on the video clip duration ($T$) used during contextual environment construction on the MMR-V benchmark.}
\label{tab:clip_duration}
\end{table}

\section{Prompts for Contextual Environment Construction}
\label{app:contextual environment construction}
In this section, we provide the prompts for factor decomposition and contextual captioning in Table~\ref{tab:prompts}.

\begin{tcolorbox}[
    breakable,
    width=\columnwidth,
    colback=white,
    colframe=gray,
    title=Prompt for factor decomposition,
    arc=0mm,
    boxrule=0.5pt,
    left=2pt, right=2pt, top=2pt, bottom=2pt,
    before skip=5pt, after skip=5pt,
    fonttitle=\bfseries\small,
]
\begin{lstlisting}[
    basicstyle=\ttfamily\footnotesize,
    numbers=none,
    tabsize=2,
    breaklines=true,
    breakatwhitespace=true,
    aboveskip=2pt,
    belowskip=2pt,
    breaklines=true,
]
// system prompt
You are a helpful assistant.

// user prompt
You are given a question about a single video.
Your first task is to synthesize a prioritized list of k key factors that are most critical to answer these questions accurately.
- Choose k adaptively within [K_MIN, K_MAX], balancing coverage and redundancy. 
- Each factor should be concise (2-6 words), actionable, and non-overlapping. 
- For each factor, provide: a short name, a category, why it matters, and what evidence to collect.

Return a single JSON object with the schema:
{
  "factor_budget": {"k": <int>, "reason": "<why this k>"},
  "selected_factors": [
    {
      "name": "<factor name>",
      "category": "entity|action|attribute| relation|spatial|temporal| counting|text_ocr|audio|other ",
      "why": "<1-line rationale>",
      "evidence_to_collect": ["<brief hints of what to look for>"]
    }
  ],
  "ignore_list": ["<optional: clearly irrelevant or misleading leads>"],

  // Optional expanded buckets to help later (may be empty)
  "categories": {
    "entities": [string],
    "actions": [string],
    "attributes": [string],
    "relations": [string],
    "spatial": [string],
    "temporal": [string],
    "counting_targets": [string],
    "text_ocr": [string],
    "audio_speech": [string],
    "negatives": [string]
  }
}

Questions:
QUESTIONS_PLACEHOLDER

Return only the JSON.
\end{lstlisting}
\end{tcolorbox}

\begin{tcolorbox}[
    breakable,
    width=\columnwidth,
    colback=white,
    colframe=gray,
    title=Prompt for question-conditioned captioning,
    arc=0mm,
    boxrule=0.5pt,
    left=2pt, right=2pt, top=2pt, bottom=2pt,
    before skip=5pt, after skip=5pt,
    fonttitle=\bfseries\small,
]
\begin{lstlisting}[
    basicstyle=\ttfamily\footnotesize,
    numbers=none,
    tabsize=2,
    breaklines=true,
    breakatwhitespace=true,
    aboveskip=2pt,
    belowskip=2pt,
]
// system prompt
You are a helpful assistant.

// user prompt
There are consecutive frames from a video. Please understand the video clip with the given transcript and the selected question-aware factors. Output JSON in the template below.

Formatting rules:
- Output must be valid JSON (no comments in the final output).
- Use double quotes for all strings. Escape inner quotes (e.g. \"text\").
- Use null for unknown/not-applicable values.
- Times must be HH:MM:SS (zero-padded), relative to the source video.

Transcript of current clip:
TRANSCRIPT_PLACEHOLDER

Selected factors (adaptive k):
PRIORS_PLACEHOLDER

Output template:
{
  "clip_start_time": "CLIP_START_TIME",                 // DO NOT MODIFY - This is the exact start time of the provided video clip
  "clip_end_time": "CLIP_END_TIME",                     // DO NOT MODIFY - This is the exact end time of the provided video clip
  "subject_registry": {
    "<subject_i>": {
      "name": "<short identity if name is unknown>",    // e.g., "man in red jacket"; use given name if explicitly provided
      "appearance": ["..."],                             // stable visual traits: clothing colors, accessories, age group, ...
      "identity": ["..."],                               // list of identity descriptions
      "first_seen": "<timestamp>"                        // HH:MM:SS when this subject first appears in the clip
    },
    "...": {}
  },
  "clip_description": "<smooth and detailed natural narration based on video frames and transcript>",

  "qa_factor_evidence": [
    {
      "factor": "<one of selected_factors.name>",      
      "category": "<its category>",                      // copy exactly from the selected factor's category
      "present": true|false,                             // true if directly supported by frames/transcript; false if absent/unclear
      "count": <int|null>,                               // only when relevant; otherwise null
      "text": "<ocr text|null>",                         // only when relevant; otherwise null
      "evidence": "<short, specific visual/transcript cue>",  // If present: e.g., 'logo on jersey'. If absent: 'Not visible' or explain occlusion.
      "time_range": [["HH:MM:SS", "HH:MM:SS"]]           // List of time intervals [start, end] in video time during which the evidence is observable; empty array if not detectable
      "notes": "<optional clarification>"
    }
  ]
}

Construction guidance:
- subject_registry: include subjects necessary to explain the selected factors; assign stable keys (e.g., "S1", "S2").
- appearance: use concise, neutral noun phrases; avoid actions or opinions.
- qa_factor_evidence: set present=false when evidence is missing or ambiguous; in that case, use null for count and text.
\end{lstlisting}
\end{tcolorbox}

\captionof{table}{ \label{tab:prompts} The prompts for factor decomposition, question-conditioned captioning in \dvd{}.}

\section{Case Study}
\label{app:case study}

In this section, we illustrate some examples of success and failure cases of \dvd{} on the MMR-V Dataset in Table~\ref{tab:case study correct standardized 4-steps trajectory}-\ref{tab:case study metaphor}. 
For clarity,  \textcolor{red}{accurate and useful} content is indicated in \textcolor{red}{red}, and \textcolor{orange}{inaccurate or irrelevant} content is indicated in \textcolor{orange}{orange}.

\label{sec:appendix}

\clearpage

\begin{tcolorbox}[
    colback=white,
    colframe=gray,
    title=Question,
    breakable,
    enhanced,
    boxrule=0.5pt,
    arc=2pt,
    top=5pt,
    bottom=5pt,
]
The shoes of the female protagonist change at 2:00 and 6:50 in the video. What do these changes represent respectively?

(A) It means that the girl is hallucinating due to the extreme pressure of space travel, and she's seeing things that aren't really there.

(B) It signifies that the space environment has some kind of magical power that can transform any object, and it just randomly changed her boots.

(C) The girl misses her father dearly and has imprinted the pattern of her childhood shoes onto the boots she wears when stepping into space, symbolizing the fulfillment of her childhood dream.

(D) It symbolizes that the girl regrets becoming an astronaut and wishes she could go back to her carefree childhood.

(E) It implies that the girl is actually in a virtual reality simulation of space travel, and the developers added this nostalgic element for effect.

(F) It means that the girl has a split personality, and the appearance of the childhood boots is a manifestation of her inner child taking over.

(G) This implies that the girl has fulfilled her childhood dream and also recalls the support and love from her father.

(H) It shows that there's a time - travel anomaly in space, and the boots are a physical sign of her being briefly transported back to her childhood.

(I) It indicates that the girl is under the influence of some alien - made mind - altering substance that makes her see her boots transform into childhood ones.

\end{tcolorbox}

\begin{tcolorbox}[
    colback=white,
    colframe=brown,
    title=Agent (Step 1),
    breakable,
    boxrule=0.5pt,
    arc=2pt,
    top=5pt,
    bottom=5pt,
]
Calling \textbf{Global Browse} with event query "What are the main events and subjects in the video regarding the girl's space journey and childhood?"
\end{tcolorbox}

\begin{tcolorbox}[
    colback=white,
    colframe=cyan!90,
    title=Tool Response (Global Browse),
    breakable,
    boxrule=0.5pt,
    arc=2pt,
    top=5pt,
    bottom=5pt,
]
\textbf{Subjects-centric Summary}:

\begin{lstlisting}[
basicstyle=\small,
frame=none, 
columns=fullflexible,
backgroundcolor=\color{white}, 
showstringspaces=false, 
escapeinside={(*@}{@*)}
]
...
 "Father": {
    "name": "Father",
    "appearance": ...,
    "identity": [
      "(*@\textcolor{red}{shoemaker}@*)",
    ],
}
\end{lstlisting}

\textbf{Event-centric Summary}:

... whose \textcolor{red}{childhood fascination with space} evolves into her eventual \textcolor{red}{realization of becoming an astronaut}. ... \textcolor{red}{The father-daughter relationship} is a central theme throughout the video...

\end{tcolorbox}

\begin{tcolorbox}[
    colback=white,
    colframe=brown,
    breakable,
    title=Agent (Step 2),
    boxrule=0.5pt,
    arc=2pt,
    top=5pt,
    bottom=5pt,
]
Calling \textbf{Query Decompose} with the query and all options.
\end{tcolorbox}

\begin{tcolorbox}[
    colback=white,
    colframe=cyan!90,
    title=Tool Response (Query Decompose),
    breakable,
    boxrule=0.5pt,
    arc=2pt,
    top=5pt,
    bottom=5pt,
]
Analysis: ... 

Decisive Factor: Positive Symbolism vs. Negative/External Cause
...

If the transformation signifies fulfillment of a meaningful dream $\longrightarrow $ (C), (G).

If the transformation is a symptom of psychological distress or external forces $\longrightarrow$ (A, B, D, E, F, H, I).
\end{tcolorbox}

\begin{tcolorbox}[
    colback=white,
    colframe=brown,
    title=Agent (Step 3),
    breakable,
    boxrule=0.5pt,
    arc=2pt,
    top=5pt,
    bottom=5pt,
]
Calling \textbf{Frame Inspect} at ["00:06:50", "00:06:59"]: "Analyze the visual details of the boots transforming into childhood boots."

\end{tcolorbox}

\begin{tcolorbox}[
    colback=white,
    colframe=cyan!90,
    title=Tool Response (Frame Inspect),
    boxrule=0.5pt,
    breakable,
    arc=2pt,
    top=5pt,
    bottom=5pt,
]
The boots feature a rocket logo, which appears both \textcolor{red}{in the childhood} scene (worn by the girl as a child) and later \textcolor{red}{in the space} - setting (as part of her astronaut gear)...The girl's smile, closed eyes in contentment during the hug with her father, ..., filled backdrop all convey a sense of \textcolor{red}{fulfillment and emotional connection}. 
\end{tcolorbox}

\begin{tcolorbox}[
    colback=white,
    colframe=brown,
    title=Agent (Step 6),
    boxrule=0.5pt,
    breakable,
    arc=2pt,
    top=5pt,
    bottom=5pt,
]
Answering: (G) $\longleftarrow$ Correct
\end{tcolorbox}
\captionof{table}{Case Study: Correct standardized 4-steps trajectory}
\label{tab:case study correct standardized 4-steps trajectory}




\clearpage

\centering
\begin{tcolorbox}[
    colback=white,
    colframe=gray,
    title=Question,
    breakable,
    enhanced,
    boxrule=0.5pt,
    arc=2pt,
    top=5pt,
    bottom=5pt,
]

Why do several people need to pass around that playing card after 3:10? \\
(A) The card contains a hidden camera used to surveil the area. \\
(B) The card carries a cryptic map to a secret exit in the building. \\
(C) The card acts as an access key for security purposes and is part of their plan to blend in with the system. \\
(D) The card signals other team members when to act by changing color. \\
(E) The act of passing the card is a diversion to confuse security guards. \\
(F) The card is being used to disable an automated security system. \\
(G) The card unlocks a vault containing confidential documents. \\
(H) The characters are betting with the card to decide who takes the next step in the plan. \\
(I) The card is a tracking device meant to monitor everyone's location in real time. \\
(J) The characters are simply playing a casual game to distract themselves. \\
(K) They are passing the chip that is attached to the back of the playing card. \\
\end{tcolorbox}

\begin{tcolorbox}[
    colback=white,
    colframe=brown,
    title=Agent (Step 1),
    breakable,
    boxrule=0.5pt,
    arc=2pt,
    top=5pt,
    bottom=5pt,
]
Calling \textbf{Global Browse} with event query "Global overview of events and main subjects in the video."
\end{tcolorbox}

\begin{tcolorbox}[
    colback=white,
    colframe=cyan!90,
    title=Tool Response (Global Browse),
    breakable,
    boxrule=0.5pt,
    arc=2pt,
    top=5pt,
    bottom=5pt,
]
\textbf{Subjects-centric Summary}:

\begin{lstlisting}[
basicstyle=\small,
frame=none, 
columns=fullflexible,
backgroundcolor=\color{white}, 
showstringspaces=false, 
escapeinside={(*@}{@*)}
]
...
 "S6": {
    "name": "man in gray suit",
    "appearance": ...,
    "identity": ...,          
}
...
 "S8": {
    "name": "woman in white lab coat",
    "appearance": ...,
    "identity": ...,                   
 }
 "S9": {
    "name": "security guard (background)",
    "appearance": ...,
    "identity": ...,
    }, 
...
}
\end{lstlisting}

\textbf{Event-centric Summary}:

This video sequence depicts a complex narrative set in a high-tech security facility, ... \textcolor{red}{Various individuals present playing card}, ...

\end{tcolorbox}

\begin{tcolorbox}[
    colback=white,
    colframe=brown,
    breakable,
    title=Agent (Step 2),
    boxrule=0.5pt,
    arc=2pt,
    top=5pt,
    bottom=5pt,
]
Calling \textbf{Query Decompose} with the query and all options.
\end{tcolorbox}

\begin{tcolorbox}[
    colback=white,
    colframe=cyan!90,
    title=Tool Response (Query Decompose),
    breakable,
    boxrule=0.5pt,
    arc=2pt,
    top=5pt,
    bottom=5pt,
]
Analysis: All options frame the card-passing as part of a deliberate, ...

Decisive Factor: Whether the card’s purpose stems from \textcolor{red}{the card’s intrinsic properties (A–J)} or \textcolor{red}{a separate device attached to it (K).}
\end{tcolorbox}

\begin{tcolorbox}[
    colback=white,
    colframe=brown,
    title=Agent (Step 3),
    breakable,
    boxrule=0.5pt,
    arc=2pt,
    top=5pt,
    bottom=5pt,
]
Calling \textbf{Evidence Retrieve} to obtain the types of available evidence.
\end{tcolorbox}

\begin{tcolorbox}[
    colback=white,
    colframe=cyan!90,
    title=Tool Response (Evidence Retrieve),
    boxrule=0.5pt,
    breakable,
    arc=2pt,
    top=5pt,
    bottom=5pt,
]

['Card function', 'Hidden device presence']

\end{tcolorbox}

\begin{tcolorbox}[
    colback=white,
    colframe=brown,
    title=Agent (Step 4),
    boxrule=0.5pt,
    breakable,
    arc=2pt,
    top=5pt,
    bottom=5pt,
]
Calling \textbf{Evidence Retrieve} with evidence type: "Hidden device presence"
\end{tcolorbox}

\begin{tcolorbox}[
    colback=white,
    colframe=cyan!90,
    title=Tool Response (Evidence Retrieve),
    boxrule=0.5pt,
    breakable,
    arc=2pt,
    top=5pt,
    bottom=5pt,
]
\begin{lstlisting}[columns=fullflexible,basicstyle=\small, frame=none, backgroundcolor=\color{white}, breaklines=true, showstringspaces=false, escapeinside={(*@}{@*)}]
...
{
  "evidence": "Hands manipulating circuit board and (*@\textcolor{red}{\mbox{small chip}}@*) under blue lighting",
  "time_range": [
    [
      "00:02:15",
      "00:02:19"
    ]
  ],
  "notes": "Electronic components visible during dark blue-lit sequence"
},
{
  "evidence": "Close-up shows a (*@\textcolor{red}{\mbox{small chip attached to the playing card}}@*)",
  "time_range": [
    [
      (*@\textcolor{orange}{"00:00:12"}@*),
      (*@\textcolor{orange}{"00:00:16"}@*)
    ]
  ],
  "notes": "Chip is the hidden device"
},
...
\end{lstlisting}
\end{tcolorbox}

\begin{tcolorbox}[
    colback=white,
    colframe=brown,
    title=Agent (Step 5),
    boxrule=0.5pt,
    breakable,
    arc=2pt,
    top=5pt,
    bottom=5pt,
]
Calling \textbf{Frame Inspect} at ["03:10", "03:15"]: "Analyze the card-passing sequence after 3:10."

\end{tcolorbox}

\begin{tcolorbox}[
    colback=white,
    colframe=cyan!90,
    title=Tool Response (Frame Inspect),
    boxrule=0.5pt,
    breakable,
    arc=2pt,
    top=5pt,
    bottom=5pt,
]
... The critical observation is that the individual holds a card (visible in one frame) and then the card is seen falling to the ground. There is no indication of the card being passed between individuals. ...
\end{tcolorbox}

\begin{tcolorbox}[
    colback=white,
    colframe=brown,
    title=Agent (Step 6),
    boxrule=0.5pt,
    breakable,
    arc=2pt,
    top=5pt,
    bottom=5pt,
]
Answering: (K) $\longleftarrow$ Correct 
\end{tcolorbox}
\captionof{table}{Case Study: Correct multiple steps trajectory}
\label{tab:case study correct multiple steps trajectory}






\clearpage

\begin{tcolorbox}[
    colback=white,
    colframe=gray,
    title=Question,
    breakable,
    enhanced,
    boxrule=0.5pt,
    arc=2pt,
    top=5pt,
    bottom=5pt,
]
How did the man at the beginning of the film manage to pull out half of his finger and then reattach it?

(A) He used a specially designed prosthetic finger that could detach and reattach.

(B) An accomplice assisted using a mirror to create the illusion of detachment and reattachment.

(C) He used a hidden compartment within his sleeve to conceal the finger temporarily.

(D) He applied a strong adhesive gel that allowed him to temporarily detach the finger.

(E) He simply bent his finger to create an optical illusion.

(F) The video was edited to create the illusion of the finger being detached and reattached.

(G) He wore a finger glove that gave the illusion of detachment.

(H) Special effects makeup was applied to create the appearance of a removable finger.

(I) He had a double-jointed finger that allowed him to do this trick.

(J) He used magnets within his hands to achieve the illusion.

\end{tcolorbox}

\begin{tcolorbox}[
    colback=white,
    colframe=brown,
    title=Agent (Step 1),
    breakable,
    boxrule=0.5pt,
    arc=2pt,
    top=5pt,
    bottom=5pt,
]
Calling \textbf{Global Browse} with event query "How did the man at the beginning of the film manage to pull out half of his finger and then reattach it?"
\end{tcolorbox}

\begin{tcolorbox}[
    colback=white,
    colframe=cyan!90,
    title=Tool Response (Global Browse),
    breakable,
    boxrule=0.5pt,
    arc=2pt,
    top=5pt,
    bottom=5pt,
]
\textbf{Subjects-centric Summary}:

\begin{lstlisting}[
basicstyle=\small,
frame=none, 
columns=fullflexible,
backgroundcolor=\color{white}, 
showstringspaces=false, 
escapeinside={(*@}{@*)}
]
...

\end{lstlisting}

\textbf{Event-centric Summary}:

...there is no explicit description of this specific finger trick occurring in any of the provided clips.

\end{tcolorbox}

\begin{tcolorbox}[
    colback=white,
    colframe=brown,
    breakable,
    title=Agent (Step 2),
    boxrule=0.5pt,
    arc=2pt,
    top=5pt,
    bottom=5pt,
]
Calling \textbf{Query Decompose} with the query and all options.
\end{tcolorbox}

\begin{tcolorbox}[
    colback=white,
    colframe=cyan!90,
    title=Tool Response (Query Decompose),
    breakable,
    boxrule=0.5pt,
    arc=2pt,
    top=5pt,
    bottom=5pt,
]

Analysis: ... 

Decisive Factor: The method must be \textcolor{red}{physically performable} during the scene (in-camera), excluding post-production editing (F).

\end{tcolorbox}

\begin{tcolorbox}[
    colback=white,
    colframe=brown,
    title=Agent (Step 3),
    boxrule=0.5pt,
    breakable,
    arc=2pt,
    top=5pt,
    bottom=5pt,
]
Calling \textbf{Frame Inspect} at [                "00:00:00", "00:00:05"]: "Analyze the man's hand movements at the beginning of the film to determine if the finger trick described in the question is physically possible or if it was created through digital editing."

\end{tcolorbox}

\begin{tcolorbox}[
    colback=white,
    colframe=cyan!90,
    title=Tool Response (Frame Inspect),
    boxrule=0.5pt,
    breakable,
    arc=2pt,
    top=5pt,
    bottom=5pt,
]
The provided video frames show the man performing \textcolor{orange}{intricate hand movements, which might be a prosthetic}. ...
\end{tcolorbox}

\begin{tcolorbox}[
    colback=white,
    colframe=brown,
    title=Agent (Step 6),
    boxrule=0.5pt,
    breakable,
    arc=2pt,
    top=5pt,
    bottom=5pt,
]
Answering: (A) $\longleftarrow$ Wrong, correct answer is (E)
\end{tcolorbox}

\captionof{table}{Case Study: Failed standardized 4-steps trajectory}
\label{tab:case study failed standardized 4-steps trajectory}




\clearpage

\begin{tcolorbox}[
    colback=white,
    colframe=gray,
    title=Question,
    breakable,
    enhanced,
    boxrule=0.5pt,
    arc=2pt,
    top=5pt,
    bottom=5pt,
]
What does the man looking into the mirror symbolize?

(A) An illusion of duality.

(B) The reflection of one's younger self and the passage of time.

(C) An ancient myth of reflection.

(D) An audience behind a glass.

(E) Superficial appearance correction.

(F) Vanity and self-admiration.

(G) A parallel universe interaction.

(H) A future self watching over.

(I) Gazing into the mirror symbolizes introspection and contemplation of one's inner self.

(J) The path to a hidden world.

(K) A moment of narcissism.
\end{tcolorbox}

\begin{tcolorbox}[
    colback=white,
    colframe=brown,
    title=Agent (Step 1),
    breakable,
    boxrule=0.5pt,
    arc=2pt,
    top=5pt,
    bottom=5pt,
]
Calling \textbf{Global Browse} with event query "What does the man looking into the mirror symbolize?"
\end{tcolorbox}

\begin{tcolorbox}[
    colback=white,
    colframe=cyan!90,
    title=Tool Response (Global Browse),
    breakable,
    boxrule=0.5pt,
    arc=2pt,
    top=5pt,
    bottom=5pt,
]
\textbf{Subjects-centric Summary}:

\begin{lstlisting}[
basicstyle=\small,
frame=none, 
columns=fullflexible,
backgroundcolor=\color{white}, 
showstringspaces=false, 
escapeinside={(*@}{@*)}
]
...

\end{lstlisting}

\textbf{Event-centric Summary}:

...The mirror serves as a metaphorical portal \textcolor{red}{connecting the adult man to his younger self}. 
This is most clearly established in the opening sequence (00:00:0.0-00:00:9.0) \textcolor{red}{where the man in the white shirt, dark vest, and striped tie turns to face the ornate mirror, and a child in a white shirt, suspenders, and dark shorts appears within the reflection}...
**The Mirror as a Dynamic Entity**: In the sequence from 00:00:30.0-00:00:39.0, the mirror frame on wheels becomes \textcolor{red}{a central element that both separates and connects the adult and child}...
In the final moments (00:01:30.0-00:01:47.0), the man walks toward the mirror, stands facing his reflection, and \textcolor{red}{maintains eye contact while the background remains dark and starry}...
the adult fully \textcolor{orange}{embracing and acknowledging his younger self} through sustained gaze and connection.

\end{tcolorbox}

\begin{tcolorbox}[
    colback=white,
    colframe=brown,
    breakable,
    title=Agent (Step 2),
    boxrule=0.5pt,
    arc=2pt,
    top=5pt,
    bottom=5pt,
]
Calling \textbf{Query Decompose} with the query and all options.
\end{tcolorbox}

\begin{tcolorbox}[
    colback=white,
    colframe=cyan!90,
    title=Tool Response (Query Decompose),
    breakable,
    boxrule=0.5pt,
    arc=2pt,
    top=5pt,
    bottom=5pt,
]
Analysis: ... 

Decisive Factor: 

If the man's gaze implies \textcolor{red}{self-examination} of thoughts/feelings: Correct answer is (I).

If the man's gaze implies vanity, time, myth, or fantasy $\longrightarrow$ (A, B, C, D, E, F, G, H, J, K).

\end{tcolorbox}

\begin{tcolorbox}[
    colback=white,
    colframe=brown,
    title=Agent (Step 3),
    boxrule=0.5pt,
    breakable,
    arc=2pt,
    top=5pt,
    bottom=5pt,
]
Calling \textbf{Frame Inspect} at [                "00:00:00", "00:00:09"]: ""What does the man looking into the mirror symbolize?"

\end{tcolorbox}

\begin{tcolorbox}[
    colback=white,
    colframe=cyan!90,
    title=Tool Response (Frame Inspect),
    boxrule=0.5pt,
    breakable,
    arc=2pt,
    top=5pt,
    bottom=5pt,
]
The man looking into the mirror symbolizes the relationship between different life stages, as the reflection of a child within the mirror creates a visual connection between the adult and child performers.
\end{tcolorbox}

\begin{tcolorbox}[
    colback=white,
    colframe=brown,
    title=Agent (Step 6),
    boxrule=0.5pt,
    breakable,
    arc=2pt,
    top=5pt,
    bottom=5pt,
]
Answering: (B) $\longleftarrow$ Wrong, correct answer is (I)
\end{tcolorbox}

\captionof{table}{Case Study: Failed trajectory in metaphor understanding}
\label{tab:case study metaphor}

\section{The Use of Large Language Models}
In this paper, the LLMs serves as a writing assistant to help polish the content.

\end{document}